\documentclass{article}

\usepackage{amsmath,amsfonts,bm}

\def\eqref#1{equation~\ref{#1}}

\def\1{\bm{1}}

\DeclareMathAlphabet{\mathsfit}{\encodingdefault}{\sfdefault}{m}{sl}
\SetMathAlphabet{\mathsfit}{bold}{\encodingdefault}{\sfdefault}{bx}{n}

\newcommand{\thmref}[1]{Theorem~\ref{#1}}
\newtheorem{theorem}{Theorem}[section]

    \usepackage[final]{neurips_2023}

\usepackage{graphicx}
\usepackage{array}
\usepackage{ifthen}
\usepackage[normalem]{ulem}
\usepackage{wrapfig}
\usepackage{caption}

\newcommand{\bx}{\mathbf{x}}
\newcommand{\by}{\mathbf{y}}

\newcommand{\btheta}{{\boldsymbol{\theta}}}
\newcommand{\reals}{{\mathbb R}}
\newcommand{\norm}[1]{\left\|#1\right\|}
\newcommand{\zero}{{\mathbf{0}}}

\usepackage[utf8]{inputenc} 
\usepackage[T1]{fontenc}    
\usepackage{hyperref}       
\usepackage{url}            
\usepackage{booktabs}       
\usepackage{amsfonts}       
\usepackage{nicefrac}       
\usepackage{microtype}      
\usepackage{xcolor}         

\hypersetup{
	colorlinks   = true, 
	urlcolor     = blue, 
	linkcolor    = blue, 
	citecolor   = blue 
}

\usepackage[capitalise]{cleveref}

\title{\vspace{-4pt}Deconstructing Data Reconstruction:\\ Multiclass, Weight Decay and General Losses\vspace{-4pt}}

\newcommand*{\affaddr}[1]{#1}
\newcommand*{\affmark}[1][*]{\textsuperscript{#1}}
\newcommand*{\affmarkt}[1][*]{\textsuperscript{~~~#1}}
\newcommand*{\asp}{\quad\quad}

\author{%
Gon Buzaglo\thanks{Equal Contribution}\affmarkt[1]\asp Niv Haim\footnotemark[1]\affmarkt[1]\asp Gilad Yehudai\affmark[1]\asp Gal Vardi\affmark[2]\\
~~\vspace{-4pt}\\ 
\textbf{ Yakir Oz\affmark[1]\asp Yaniv Nikankin\affmark[1]\asp Michal Irani\affmark[1]}\\ 
~~\vspace{-4pt}\\
\affaddr{\affmark[1]Weizmann Institute of Science, Rehovot, Israel}\\
\affaddr{\affmark[2]TTI-Chicago and the Hebrew University of Jerusalem} \vspace{-10pt}
}

\newdimen\mywidth

\begin{document}

  \mywidth=\dimexpr
    \linewidth
    - 20\tabcolsep
  \relax

\maketitle

\begin{abstract}\vspace{-4pt}
Memorization of training data is an active research area, yet our understanding of the inner workings of neural networks is still in its infancy. Recently, \cite{haim2022reconstructing} proposed a scheme to reconstruct training samples from multilayer perceptron binary classifiers, effectively demonstrating that a large portion of training samples are encoded in the parameters of such networks. In this work, we extend their findings in several directions, including reconstruction from multiclass and convolutional neural networks. We derive a more general reconstruction scheme which is applicable to a wider range of loss functions such as regression losses. Moreover, we study the various factors that contribute to networks' susceptibility to such reconstruction schemes. Intriguingly, we observe that using weight decay during training increases reconstructability both in terms of quantity and quality. Additionally, we examine the influence of the number of neurons relative to the number of training samples on the reconstructability.
\newline Code: \url{https://github.com/gonbuzaglo/decoreco}
\vspace{-4pt}
\end{abstract}

\section{Introduction}
\vspace{-2pt}

Neural networks are known to memorize training data despite their ability to generalize well to unseen test data \citep{zhang2021understanding, feldman2020does}. This phenomenon was observed in both supervised settings~\citep{haim2022reconstructing, balle2022reconstructing,loo2023dataset} and in generative models~\citep{carlini2019secret,carlini2021extracting,carlini2023extracting}. These works shed an interesting light on generalization, memorization and explainability of neural networks, while also posing a potential privacy risk. 

Current reconstruction schemes from trained neural networks are still very limited and often rely on unrealistic assumptions, or operate within restricted settings. For instance, \cite{balle2022reconstructing} propose a reconstruction scheme based on the assumption of having complete knowledge of the training set, except for a single sample. \cite{loo2023dataset} suggest a scheme which operates under the NTK regime~\citep{jacot2018neural}, and assumes knowledge of the full set of parameters at initialization. 
Reconstruction schemes for unsupervised settings are specifically tailored for generative models and are not applicable for classifiers or other supervised tasks.

Recently, \cite{haim2022reconstructing} proposed a reconstruction scheme from feed-forward neural networks under logistic or exponential loss for binary classification tasks. Their scheme requires only knowledge of the trained parameters, and relies on theoretical results about the implicit bias of neural networks towards solutions of the maximum margin problem \citep{lyu2019gradient, ji2020directional}. Namely, neural networks are biased toward KKT points of the max-margin problem (see \thmref{thm:known KKT}). By utilizing the set of conditions that KKT points satisfy,
they devise a novel loss function that allows for reconstruction of actual training samples. They demonstrate reconstruction from models trained on common image datasets (CIFAR10~\citep{krizhevsky2009learning} and MNIST~\citep{lecun2010mnist}).

In this work, we expand the scope of neural networks for which we have evidence of successful sample memorization, by demonstrating sample reconstruction. Our contributions are as follows:

\begin{itemize}
    \item We extend the reconstruction scheme of \cite{haim2022reconstructing} to a multiclass setting (\cref{fig:teaser}). This extension utilizes the implicit bias result from \cite{lyu2019gradient} to multiclass training. We analyse the effects of the number of classes on reconstructability, and show that models become more susceptible to sample reconstruction as the number of classes increases.
    \item We devise a reconstruction scheme that applies for general loss functions, assuming that the model is trained with weight decay. We demonstrate reconstruction from models trained on regression losses.
    \item We investigate the effects of weight decay and show that for certain values, weight decay increases the vulnerability of models to sample reconstruction.
    Specifically, it allows us to reconstruct training samples from a convolutional network, while \citet{haim2022reconstructing} only handled MLPs.
    \item We analyse the intricate relation between the number of samples and the number of parameters in the trained model, and their effect on reconstrctability. We also demonstrate successful reconstruction from a model trained on $5$,$000$ samples, surpassing previous results that focused on models trained on up to $1$,$000$ samples.
\end{itemize}

\section{Related Works}

\paragraph{Memorization and Samples Reconstruction.} There is no consensus on the definition of the term ``memorization'' and different works study this from different perspectives. In ML theory, memorization usually refers to label (or, model's output) memorization~\citep{zhang2016understanding, arpit2017closer, feldman2020does, feldman2020neural, brown2021memorization}, namely, fitting the training set. Memorization in the \textit{input} domain is harder to show, because in order to demonstrate its occurrence one has to reconstruct samples from the model. \cite{balle2022reconstructing} demonstrated reconstruction of one training sample, assuming knowledge of all other training samples and \cite{haim2022reconstructing} demonstrated reconstruction of a substantial portion of training samples from a neural network classifier. \cite{loo2023dataset} extend their work to networks trained under the NTK regime \citep{jacot2018neural} and explore the relationship to dataset distillation. 
Several works have also studied memorization and samples reconstruction in generative models like autoencoders~\citep{erhan2009visualizing,radhakrishnan2018memorization}, large language models~\citep{carlini2021extracting,carlini2019secret,carlini2022quantifying} and diffusion-based image generators~\citep{carlini2023extracting,somepalli2022diffusion,gandikota2023erasing, kumari2023conceptablation}.

\paragraph{Inverting Classifiers.}
Optimizing a model's input as to minimize a class score is the common approach for neural network visualization~\citep{mahendran2015understanding}. It usually involves using input regularization~\citep{mordvintsev2015inceptionism,ghiasi2022plug}, GAN prior~\citep{nguyen2016synthesizing,nguyen2017plug} or knowledge of batch statistics~\citep{yin2020dreaming}. \cite{fredrikson2015model, yang2019neural} showed reconstruction of training samples using similar approach, however these methods are limited to classifiers trained with only a few samples per class. Reconstruction from a federated-learning setup~\citep{zhu2019deep,he2019model,hitaj2017deep,geiping2020inverting,yin2021see,huang2021evaluating,wen2022fishing} involve attacks that assume knowledge of training samples' gradients (see also \cite{wang2023reconstructing} for a theoretical analysis). In this work we do not assume any knowledge on the training data and do not use any priors other than assuming bounded inputs.

\section{Preliminaries}
\label{sec:preliminaries}

In this section, we provide an overview of the fundamental concepts and techniques required to understand the remainder of the paper, focusing on the fundamentals laid by \cite{haim2022reconstructing} for reconstructing training samples from trained neural networks.

\paragraph{Theoretical Framework.} \cite{haim2022reconstructing} builds on the theory of implicit bias of gradient descent. Neural networks are commonly trained using gradient methods, and when large enough, they are expected to fit the training data well. However, it is empirically known that these models converge to solutions that also generalize well to unseen data, despite the risk of overfitting. Several works pointed to this ``\emph{implicit bias}" of gradient methods as a possible explanation. \cite{soudry2018implicit} showed that linear classifiers trained with gradient descent on the logistic loss converge to the same solution as that of a hard-SVM, meaning that they maximize the margins. This result was later extended to non-linear and homogeneous neural networks by \cite{lyu2019gradient,ji2020directional}:

\begin{theorem}[Paraphrased from \cite{lyu2019gradient,ji2020directional}]
\label{thm:known KKT}
	Let $\Phi(\btheta;\cdot)$ be a homogeneous \footnote{A classifier $\Phi$ is homogeneous w.r.t to $\btheta$ if there exists $L\in \reals$ s.t. $\forall c\in \reals,\bx: \phi(\bx; c \btheta)=c^L \phi(\bx;\btheta)$}
	ReLU neural network. Consider minimizing 
	the logistic loss over a binary classification dataset $ \{(\bx_i,y_i)\}_{i=1}^n$ using gradient flow. Assume that there exists time $t_0$ 
    where the network classifies all the samples correctly.
	Then, gradient flow converges in direction to a first order stationary point (KKT point) of the following maximum-margin problem:
\begin{equation}
\label{eq:optimization problem}
	\min_{\btheta} \frac{1}{2} \norm{\btheta}^2 \;\;\;\; \text{s.t. } \;\;\; \forall i \in [n] \;\; y_i \Phi(\btheta; \bx_i) \geq 1~.
\end{equation}
\end{theorem}

A KKT point of \cref{eq:optimization problem} is characterized by the following set of conditions:
\begin{align}
    &\forall j \in [p], \;\;  \btheta_j - \sum_{i=1}^n  \lambda_i \nabla_{\btheta_j} \left[ y_i \Phi(\btheta; \bx_i) \right] =0~ &\text{(stationarity)}\label{eq:stationary}\\
    &\forall i \in [n], \;\;~  y_i \Phi(\btheta; \bx_i) \geq 1 &\text{(primal feasibility)}\label{eq:prim feas} \\
    &\forall i \in [n], \;\;~  \lambda_i \geq 0 &\text{(dual feasibility)}\label{eq:dual feas}\\
    &\forall i \in [n], \;\;~  \lambda_i = 0 ~ \text{if}~  y_i \Phi(\btheta; \bx_i)  \neq 1 & \text{(complementary slackness)}\label{eq:comp slack}
\end{align}

\paragraph{Reconstruction Algorithm.} \cite{haim2022reconstructing} demonstrated reconstructing samples from the training set of such classifiers by devising a reconstruction loss. Given a trained classifier $\Phi(\bx;\btheta)$, they initialize a set of $\{ \left( \bx_i,y_i \right) \}_{i=1}^m$ and $\{ \lambda_i \}_{i=1}^m$, and optimize $\bx_i,\lambda_i$ to minimize the following loss function:

\begin{equation}
    \label{eq:binary_rec_loss}
    L = \underbrace{ \Vert \btheta - \sum_{i=1}^m  \lambda_i \nabla_{\btheta_j} \left[ y_i \Phi(\btheta; \bx_i) \right] \Vert }_{L_\text{stationary}} + \underbrace{ \sum_{i=1}^m \max{ \{ -\lambda, -\lambda_\text{min} \} } }_{L_{\lambda}} + L_\text{prior} 
\end{equation}

Where $L_\text{prior}$ is simply bounding each pixel value at $\left[-1,1\right]$ \footnote{Formally:  
$L_\text{prior}=\sum_{i=1}^m \sum_{k=1}^d \max \{\max \{\bx_{i,k}-1,0 \}, \max \{ -\bx_{i,k}-1,0 \} \}$.}.
The number of training samples $n$ is unknown. However, setting $m>2n$, where $\{ y_i \}$ are set in a balanced manner allows reconstructing samples with any label distribution. The $\bx_i$'s are initialized from the Gaussian distribution $\mathcal{N}(0, \sigma_x^2 \mathbb{I})$, and 
$\lambda_\text{min},\sigma_x$ are hyperparameters.
We note that the homogeneity condition from \thmref{thm:known KKT} is not necessarily a practical limitation of this reconstruction scheme, as already in \cite{haim2022reconstructing} they show reconstructions from a non-homogeneous network.

\paragraph{Analysing and Summarizing Results.}
The optimization in \cref{eq:binary_rec_loss} is executed $k$ times (for different hyperparameters) and results in $km$ outputs ($\{ \hat{\bx}_i \}_{i=1}^{km}$) that we term \textit{candidates}, as they are candidates to be reconstructed training samples. To quantify the success of the reconstruction process, each training sample is matched with its nearest-neighbour from the $km$ candidates. The ``quality'' of reconstruction is then measured using SSIM~\cite{wang2004image} (see full details in~\cref{sec:app_analysis_details}).

An important corollary of the set of KKT conditions \cref{eq:stationary,eq:dual feas,eq:prim feas,eq:comp slack} is that the parameters of the trained model only depend on gradients of samples that are closest to the decision boundary, the so-called "margin-samples" (see end of section 3.2 in \cite{haim2022reconstructing}). Therefore, a good visual summary for analysing reconstruction from a trained model is by plotting the reconstruction quality (SSIM) against the distance from the decision boundary ($\lvert\Phi(\bx_i)\rvert$). We also utilize such visualizations. 

\paragraph{Assessing the Quality of Reconstructed Samples.}

Determining whether a candidate is a correct match for some training sample is as hard as finding a good image similarity metric. No synthetic metric such as SSIM or L2 norm can perfectly align with human perception. Perceptual similarity metrics (e.g., LPIPS~\citep{zhang2018perceptual}) build on top of pre-trained classifiers trained on Imagenet~\citep{deng2009imagenet}, and are not effective for the image resolution in this work (up to 32x32 pixels). We have observed heuristically that candidates with SSIM score higher than about $0.4$ are indeed visually similar to their nearest neighbor training sample. Hence, in this work we say that a certain candidate is a \textit{"good reconstruction"} if its SSIM score with its nearest neighbor is at least $0.4$. Also see discussion in~\cref{sec:ssim_0.4}.

\section{Reconstructing Data from Multi-Class Classifiers}
\label{sec:multiclass}

\mbox{We demonstrate that training set reconstruction can be extended to multi-class classification tasks.}

\begin{figure}[ht]
\centering
\begin{tabular}{%
    p{.1\mywidth}%
    p{.1\mywidth}%
    p{.1\mywidth}%
    p{.1\mywidth}%
    p{.1\mywidth}%
    p{.1\mywidth}%
    p{.1\mywidth}%
    p{.1\mywidth}%
    p{.1\mywidth}%
    p{.1\mywidth}%
}
\hfill Plane & \hfill Car & \hfill Bird & \hfill Cat & \hfill Deer & \hfill Dog & \hfill Frog & \hfill Horse & \hfill Ship & \hfill Truck \\
     \multicolumn{10}{l}{\includegraphics[width=\linewidth]{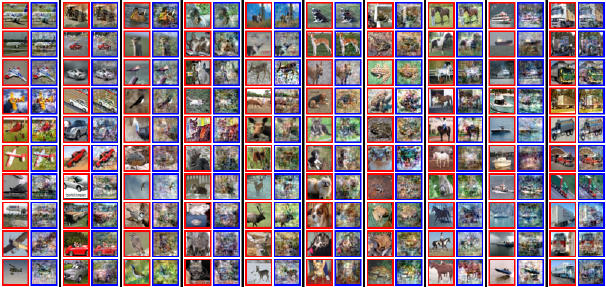}} 
\end{tabular}
\caption{Reconstructed training samples from a multi-class MLP classifier that was trained on $500$ CIFAR10 images. Each column corresponds to one class and shows the $10$ training samples (\textcolor{red}{\textit{red}}) that were best reconstructed from this class, along with their reconstructed result (\textcolor{blue}{\textit{blue}}).}
\label{fig:teaser}
\end{figure}

\subsection{Theory}
The extension of the implicit bias of homogeneous neural networks to the multi-class settings is discussed in \cite{lyu2019gradient} (Appendix G): Let $S = \{(\bx_i,y_i)\}_{i=1}^n \subseteq \reals^d \times [C]$ be a multi-class classification training set where $C\in\mathbb{N}$ is any number of classes, and $[C]= \{1,\dots,C\}$. Let $\Phi(\btheta;\cdot):\reals^d \to \reals^C$ be a homogeneous neural network parameterized by $\btheta \in \reals^p$. We denote the $j$-th output of $\Phi$ on an input $\bx$ as $\Phi_j(\btheta;\bx) \in \reals$. Consider minimizing the standard cross-entropy loss and assume that after some number of iterations the model correctly classifies all the training examples. Then, gradient flow will converge to a KKT point of the following maximum-margin problem:

\begin{equation}
\label{eq:mult_optimization problem}
	\min_{\btheta} \frac{1}{2} \norm{\btheta}^2 \;\;\;\; \text{s.t. } \;\;\; \Phi_{y_i}(\btheta; \bx_i) - \Phi_{j}(\btheta; \bx_i) \geq 1 ~~~ \forall i \in [n], \forall j \in [C] \setminus \{y_i\} \;\; ~.
\end{equation}

A KKT point of the above optimization problem is characterized by the following set of conditions:
\begin{align}
    &\btheta - \sum_{i=1}^n \sum_{j\ne{y_i}}^c\lambda_{i,j}  \nabla_{\btheta}  ( \Phi_{y_i}(\btheta; \bx_i) - \Phi_{j}(\btheta; \bx_i) ) = \zero~ \label{eq:mult_stationary}\\
    &\forall i \in [n], \forall j \in [C] \setminus \{y_i\}: \;\;  \Phi_{y_i}(\btheta; \bx_i) - \Phi_{j}(\btheta; \bx_i) \geq 1 \label{eq:mult_prim feas} \\
    &\forall i \in [n], \forall j \in [C] \setminus \{y_i\}: \;\;  \lambda_{i,j} \geq 0 \label{eq:mult_dual feas}\\
    &\forall i \in [n], \forall j \in [C] \setminus \{y_i\}: \;\;  \lambda_{i,j}= 0 ~ \text{if}~  \Phi_{y_i}(\btheta; \bx_i) - \Phi_{j}(\btheta; \bx_i) \neq 1 \label{eq:mult_comp slack}
\end{align}

A straightforward extension of a reconstruction loss for a multi-class model that converged to the conditions above would be to minimize the norm of the left-hand-side (LHS) of condition~\cref{eq:mult_stationary} (namely,  optimize over $\{\bx_i\}_{i=1}^m$ and $\{\lambda_{i,j}\}_{i \in [n], j\in [C] \setminus y_i}$ where $m$ is a hyperparameter). However, this straightforward extension failed to successfully reconstruct samples.
We therefore propose the following equivalent formulation.

Note that from \cref{eq:mult_prim feas,eq:mult_comp slack}, most $\lambda_{i,j}$ zero out:  the distance of a sample $\bx_i$ to its nearest decision boundary, \mbox{$\Phi_{y_i} - \max_{j\ne{y_i}}\Phi_j$}, is usually achieved for a single class $j$ and therefore (from \cref{eq:mult_comp slack}) in this case at most one $\lambda_{i,j}$ will be non-zero. For some samples $\bx_i$ it is also possible that all $\lambda_{i,j}$ will vanish. Following this observation, we define the following loss that only considers the distance from the decision boundary:

\begin{equation}
    L_\text{multiclass}(\bx_1,...,\bx_m,\lambda_1,...,\lambda_m)=\left\| \btheta - \sum_{i=1}^{m}\lambda_i\ \nabla_\btheta[\Phi_{y_i}(\bx_i;\btheta)-\max_{j\ne{y_i}}\Phi_j(\bx_i;\btheta)]\right\|_2^2
\label{eq:multiclass_rec_loss}
\end{equation}

\cref{eq:multiclass_rec_loss} implicitly includes~\cref{eq:mult_comp slack} into the summation in~\cref{eq:mult_stationary},  thereby significantly reducing the number of summands and simplifying the overall optimization problem. 


While the straightforward extension failed to successfully reconstruct samples, solving \cref{eq:multiclass_rec_loss} enabled reconstruction from multiclass models (see~\cref{fig:teaser} and results below). 
We attribute this success to the large reduction in the number of optimization variables, which simplifies the overall optimization problem ($n$ variables in \cref{eq:multiclass_rec_loss} compared to $C \cdot n$ variables in the straightforward case, which is significant for large number of classes $C$).


We also use the same $L_{\lambda}$ and $L_\text{prior}$ as in \cref{eq:binary_rec_loss}, and set $\{y_i\}$ in a balanced manner (uniformly on all classes). While setting $m= C\cdot n$ allows reconstructing any label distribution, in our experiments we focus on models trained on balanced training sets, and use $m=2n$ which works sufficiently well. An intuitive way to understand the extension of the binary reconstruction loss \cref{eq:binary_rec_loss} to multi-class reconstruction \cref{eq:multiclass_rec_loss} is that the only difference is the definition of the \textit{distance to nearest boundary}, which is the term inside the square brackets in both equations.

\subsection{Results}

We compare between reconstruction from binary classifiers, as studied in \cite{haim2022reconstructing}, and reconstruction from multi-class classifiers by using the novel loss function \cref{eq:multiclass_rec_loss}. We conduct the following experiment: we train an MLP classifier with architecture \mbox{$D$-$1000$-$1000$-$C$} on samples from the CIFAR10 \citep{krizhevsky2009learning} dataset. The model is trained to minimize the cross-entropy loss with full-batch gradient descent,
once with two classes ($250$ samples per class) and once for the full $10$ classes ($50$ samples per class).
Both models train on the same amount of samples ($500$).
The test set accuracy of the models is $77\%$/$32\%$ respectively, which is far from random ($50\%$/$10\%$ resp.). See implementation details in~\cref{sec:app_implemetation}.

\begin{figure}[htbp]
    \includegraphics[width=\textwidth]{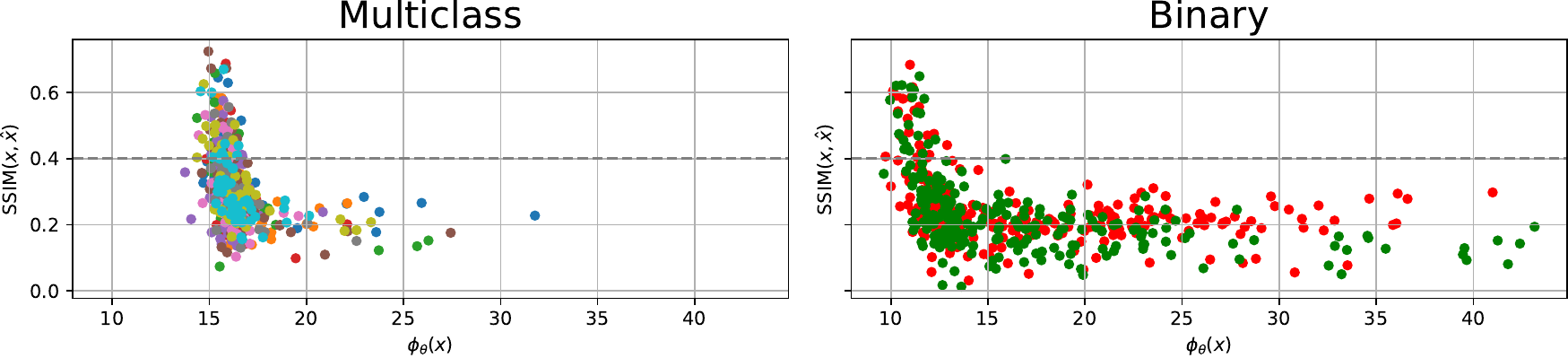}
    \caption{Multi-class classifiers are more vulnerable to training-set reconstruction. For a training set of size $500$, a multi-class model (\textit{left}) yields $101$ reconstructed samples with good quality (SSIM$>$$0.4$), compared to $40$ in a binary classification model (\textit{right}).}
    \label{fig:binary_vs_multiclass}
    \centering
\end{figure}

To quantify the quality of our reconstructed samples, for each sample in the original training set we search for its nearest neighbour in the reconstructed images and measure the similarity using SSIM~\citep{wang2004image} (higher SSIM means better reconstruction). 
In \cref{fig:binary_vs_multiclass} we plot the quality of reconstruction (in terms of SSIM) against the distance of the sample from the decision boundary $\Phi_{y_i}(\bx_i;\btheta)-\max_{j\ne{y_i}}\Phi_j(\bx_i;\btheta)$.  As seen, a multi-class classifier yields much more samples that are vulnerable to being reconstructed.

\begin{figure}[!hb]
    \begin{tabular}{l}
         \includegraphics[width=\textwidth]{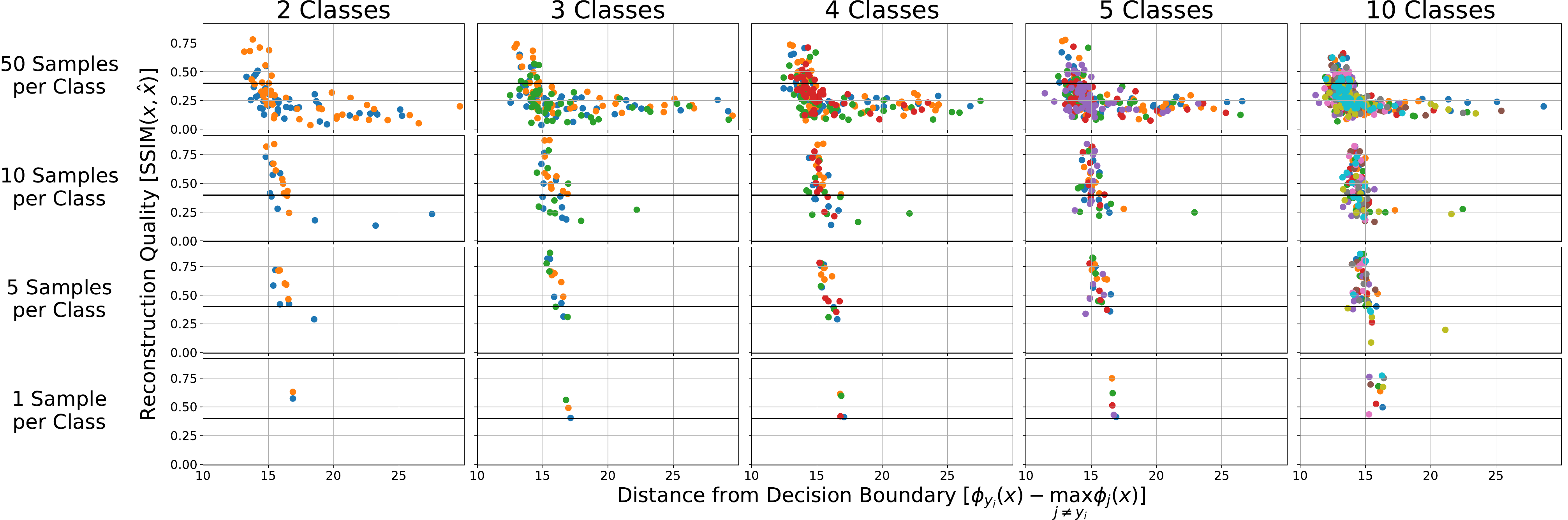}  \\
         \multicolumn{1}{l}{(a) Analysing the relation between number of classes and number of samples per class} \\
         \multicolumn{1}{l}{\hspace{0.4cm} in reconstruction from multiclass Classifiers.} \\
         \includegraphics[width=\textwidth]{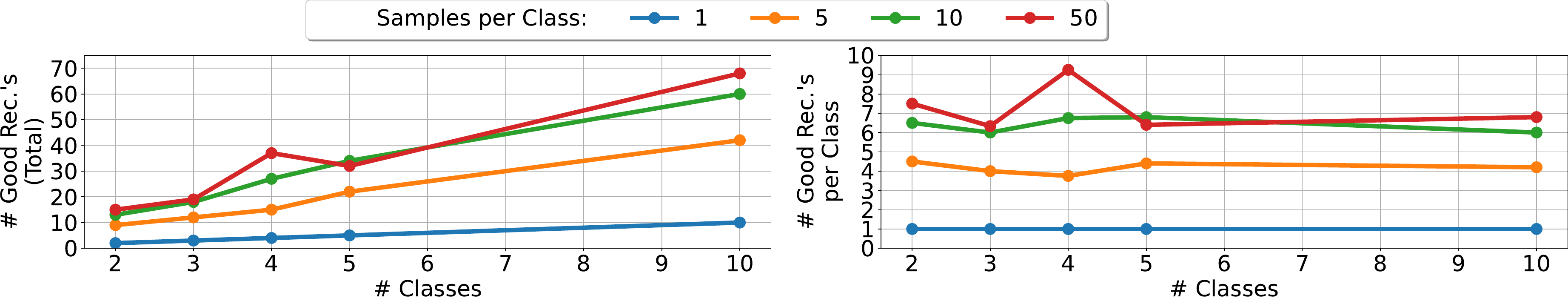} \\
         (b) Number of "good" reconstructions increases with number of classes and the samples per class
    \end{tabular}
    \caption{Evaluating the effect of multiple classes on the ability to reconstruct. We show reconstructions from models trained with different numbers of classes and different numbers of samples per class. As seen, multiple classes result in more reconstructed samples.}
    \label{fig:c_vs_dpc}
    \centering
\end{figure}

We examine the relation between the ability to reconstruct from a model and the number of classes on which it was trained. Comparing between two models trained on different number of classes is not immediately clear, since we want to isolate the effect of the number of classes from the size of the dataset (it was observed by \citet{haim2022reconstructing} that the number of reconstructed samples decreases as the total size of the training set increases). We therefore train models on training sets with varying number of classes ($C \in \{2,3,4,5,10\}$) and varying number of samples per class ($1,5,10,50$). The results are visualized in \cref{fig:c_vs_dpc}a. As seen, for models with same number of samples per class, the ability to reconstruct \emph{increases} with the number of classes, even though the total size of the training set is larger. This further validates our hypothesis that the more classes, the more samples are vulnerable to reconstruction (also see \cref{sec:fixed_dataset_size}).

Another way to validate this hypothesis is by showing the dependency between the number of classes and the number of ``good'' reconstructions (SSIM$>$$0.4$) -- shown in~\cref{fig:c_vs_dpc}b. As can be seen, training on multiple classes yields more samples that are vulnerable to reconstruction.
An intuitive explanation, is that multi-class classifiers have more ``margin-samples". Since margin-samples are more vulnerable to reconstruction, this results in more samples being reconstructed from the model.

\section{Data Reconstruction with General Loss Functions}
\label{sec:general_losses}

We demonstrate that data reconstruction can be generalized to a larger family of loss functions. 
\cite{haim2022reconstructing} and \cref{sec:multiclass}
only considered a reconstruction scheme based on the implicit bias of gradient methods trained with cross-entropy loss. For other loss functions, such as the square loss, a precise characterization of the implicit bias in nonlinear networks does not exist \citep{vardi2021implicit}. Hence, we establish a reconstruction scheme for networks trained with explicit regularization, i.e., with weight decay. 
We show that as long as the training involves a weight-decay term, we can derive a reconstruction objective that is very similar to the previous objectives in \cref{eq:binary_rec_loss,eq:multiclass_rec_loss}. 

\subsection{Theory} \label{sec:general loss theory}
Let $\ell(\Phi(\bx_i;\btheta), y_i)$ be a loss function that gets as input the predicted output of the model $\Phi$ (parametrized by $\btheta$) on an input sample $\bx_i$, and its corresponding label $y_i$. The total regularized loss:

\begin{equation}
    \label{eq:general_loss_with_wd}
    \mathcal{L} = \sum_{i=1}^n \ell(\Phi(\bx_i;\btheta), y_i) + \lambda_{\text{WD}} \frac{1}{2} \Vert \btheta \Vert^2 ~~~~.
\end{equation}

Assuming convergence ($\nabla_{\btheta} \mathcal{L} = 0$), the parameters should satisfy the following :
\begin{equation} \label{eq:reconstruction eq general loss}
    \btheta - \sum_{i=1}^n \ell'_i~ \nabla_{\btheta} \Phi(\bx_i;\btheta) = 0
\end{equation}

where $\ell'_i=\frac{1}{\lambda_{WD}} \frac{\partial \ell(\Phi(\bx_i;\btheta),y_i)}{\partial \Phi(\bx_i;\btheta)}$. This form (which is similar to the condition in \cref{eq:stationary}), 
 allows us to define a generalized reconstruction loss for models trained with a weight-decay term:

\begin{equation}
    \label{eq:general_loss}
    L_{rec}(\bx_1,...,\bx_m,\lambda_1,...,\lambda_m) = \Vert \btheta - \sum_{i=1}^n \lambda_i \nabla_{\btheta} \Phi(\bx_i;\btheta) \Vert_2^2
\end{equation}

As before, we also include the same $L_\text{prior}$ as in \cref{sec:preliminaries}. It is straightforward to see that $L_{rec}$ is a generalization of the reconstruction loss in \cref{eq:binary_rec_loss} ($y_i$ could be incorporated into the $\lambda_i$ term).

\subsection{Results and Analysis}

\begin{figure}[htbp]
    \includegraphics[width=\textwidth]{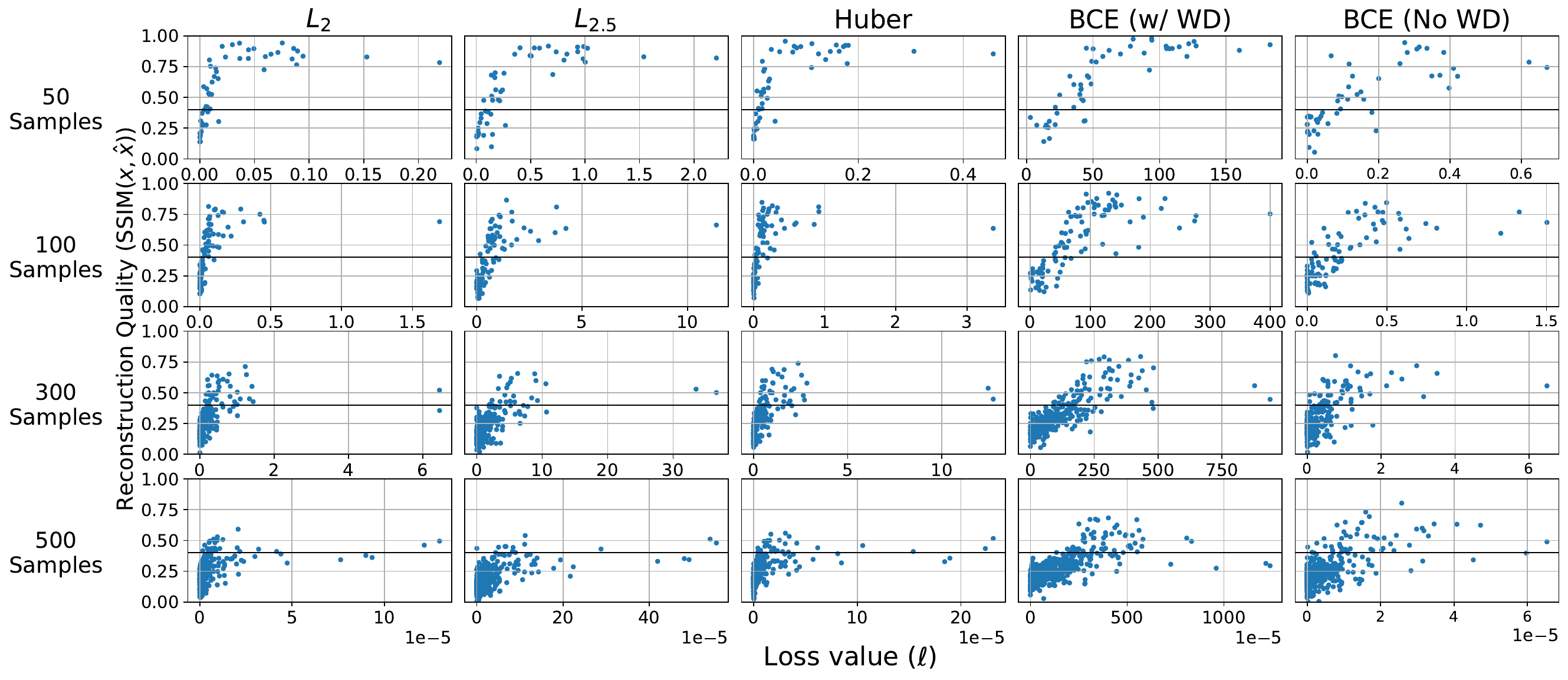}
    \caption{\textbf{Reconstruction from general losses} (column) for various training set sizes (row), using~\cref{eq:general_loss}. ``Harder'' samples (with higher loss) are easier to reconstruct.}
\label{fig:general_losses}
    \centering
\end{figure}

We validate the above theoretical analysis by demonstrating reconstruction from models trained on other losses than the ones shown in \cref{sec:multiclass} and \cite{haim2022reconstructing}. We use the same dataset as in the classification tasks -- images from CIFAR10 dataset with binary labels of $\{-1,1\}$. The only difference is replacing the classification binary cross-entropy loss with regression losses (e.g., MSE).

In classification tasks, we analyzed the results by plotting the reconstruction quality (SSIM) against the sample's distance from the decision boundary (see~\cref{sec:preliminaries}). This showed that reconstruction is only feasible for margin-samples. However, in regression tasks, margin and decision boundary lack specific meaning. We propose an alternative analysis approach -- note that smaller distance from the margin results in higher loss for binary cross-entropy. Intuitively, margin-samples are the most challenging to classify (as reflected by the loss function). Therefore, for regression tasks, we analyze the results by plotting the reconstruction quality against the loss (per training sample).

In \cref{fig:general_losses} we show results for reconstructions from models trained with MSE, $L_{2.5}$ loss ($\ell=\vert \Phi(\bx;\btheta) - y \vert^p$ for $p$=$2$,$2.5$ respectively) and Huber loss~\citep{huber1992robust}.
The reconstruction scheme in~\cref{eq:general_loss} is the same for all cases, and is invariant to the loss used during training. \cref{fig:general_losses} highlights two important observations: first, the reconstruction scheme in~\cref{eq:general_loss} succeeds in reconstructing large portions of the training set from models trained with regression losses, as noted from the high quality (SSIM) of the samples. Second, by plotting quality against the loss, one sees that ``challenging'' samples (with high loss) are easier to reconstruct. Also note that the analysis works for classification losses, namely BCE with or without weight-decay in~\cref{fig:general_losses}). For more results see~\cref{sec:app_general_losses}.

\section{On the Different Factors that Affect Reconstructability}

Our goal is to gain a deeper understanding of the factors behind models' vulnerability to reconstruction schemes. In this section, we present several analyses that shed light on several important factors.

\subsection{The Role of Weight Decay in Data Reconstruction}
\label{sec:weight_decay}

\cite{haim2022reconstructing} assumed MLP models whose first fully-connected layer was initialized with small (non-standard) weights. Models with standard initialization (e.g., \cite{he2015delving,glorot2010understanding})
did not yield reconstructed samples. In contrast, the MLPs reconstructed in \cite{haim2022reconstructing} were initialized with an extremely small variance in the first layer.
Set to better understand this drawback, we observed that incorporating weight-decay during training, not only enabled samples reconstruction in models with standard initialization, but often increase the reconstructability of training samples.

\noindent%
\begin{minipage}{\linewidth}%
\makebox[\linewidth]{%
    \begin{tabular}{ccc}
         \includegraphics[keepaspectratio=true,width=.32\textwidth]{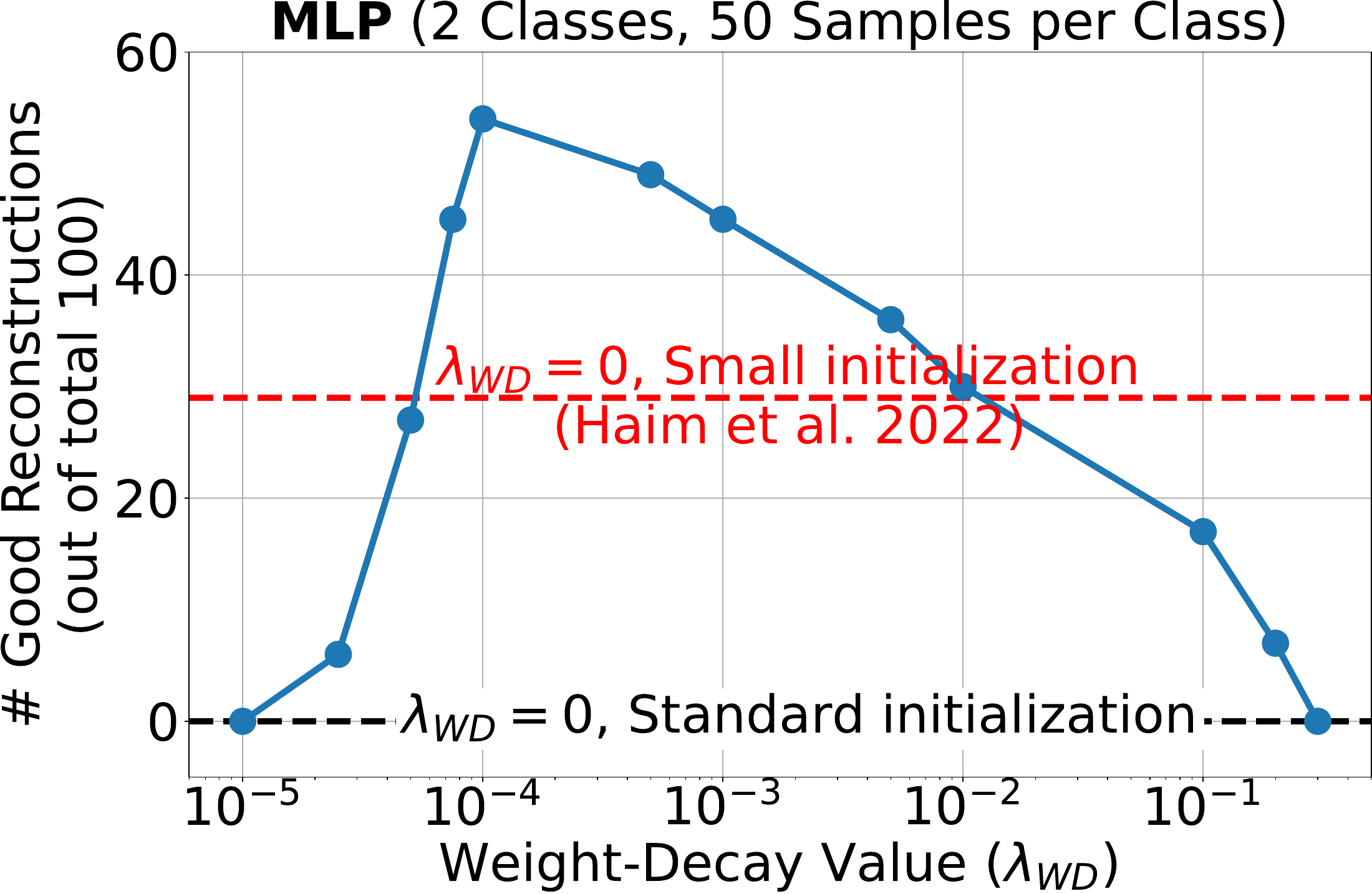} &  
         \includegraphics[keepaspectratio=true,width=.32\textwidth]{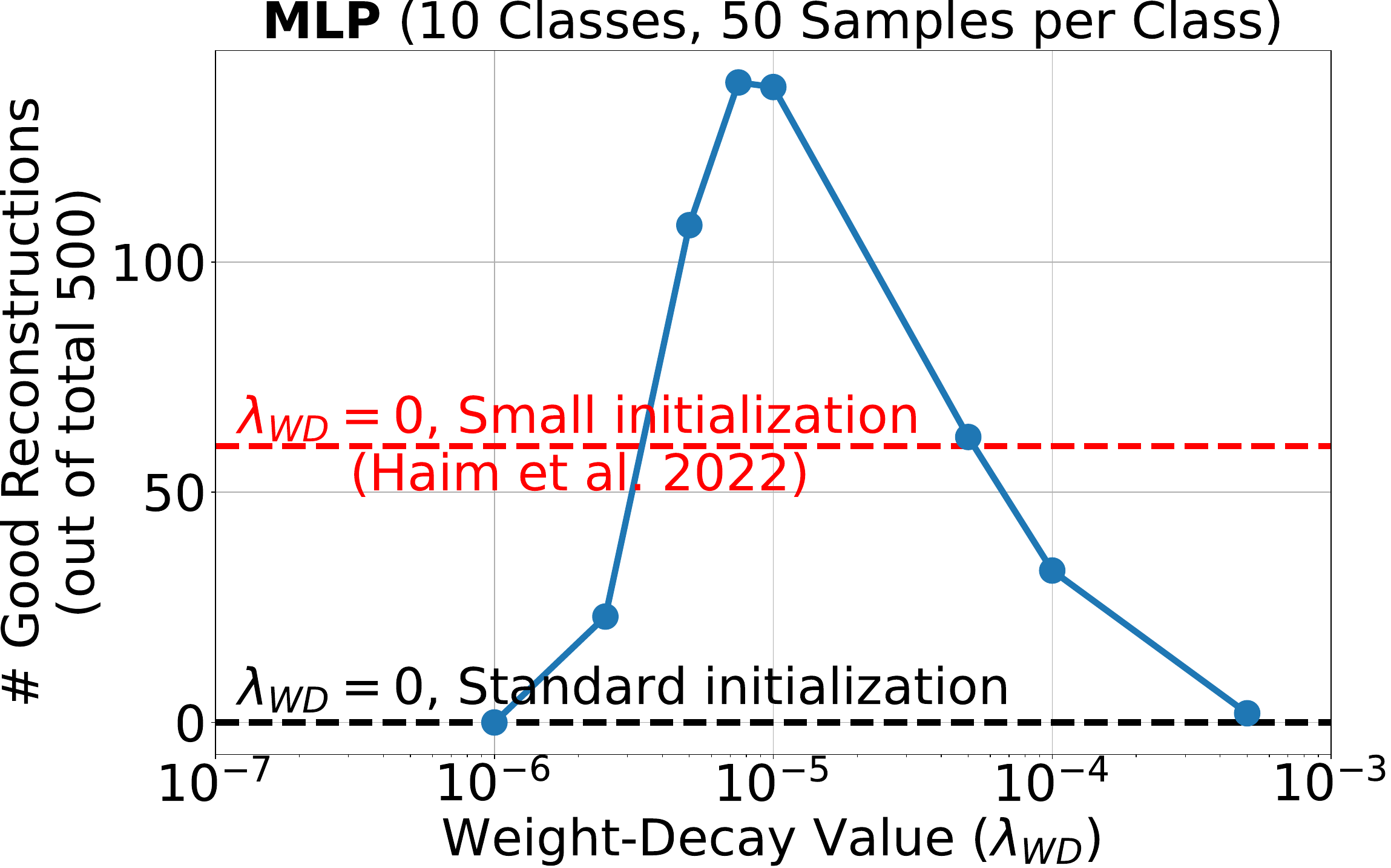} &
         \includegraphics[keepaspectratio=true,width=.32\textwidth]{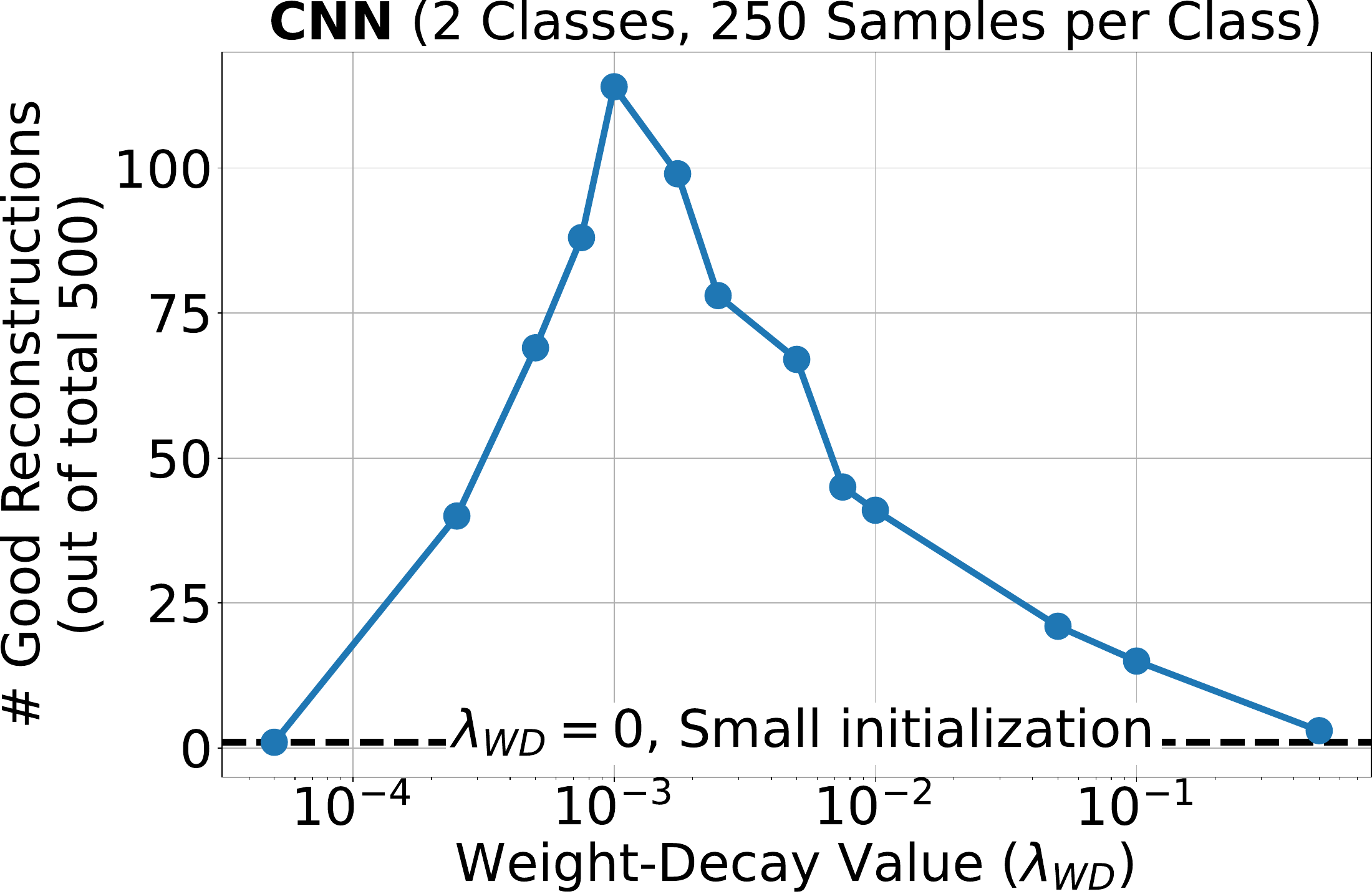} 
         \\ (a) & (b) & (c)
    \end{tabular}
  }
\captionof{figure}{Using weight-decay during training increases vulnerability to sample reconstruction}\label{fig:weight_decay}%
\end{minipage}

In \cref{fig:weight_decay}a-b we show the number of good reconstructions for different choices of the value of the weight decay ($\lambda_{\text{WD}}$), for MLP classifiers trained on $C$=$2$,$10$ classes and  $50$ samples per class (\cref{fig:weight_decay} a, b resp.). We add two baselines trained \emph{without} weight-decay: model trained with standard initialization (black) and model with small-initialized first-layer (red). 
See how for some values of weight-decay, the reconstructability is \emph{significantly higher} than what was observed for models with non-standard initialization. By examining the training samples' distance to the boundary, one observes that using weight-decay results in more margin-samples which are empirically more vulnerable to reconstruction (see full details in \cref{sec:app_weight_decay}). 

We now give an intuitive theoretical explanation to the role of weight decay in data reconstruction.
\thmref{thm:known KKT} is used to devise the reconstruction loss in \cref{eq:binary_rec_loss}, which is based on the directional convergence to a KKT point of the max-margin problem. However, this directional convergence occurs asymptotically as the time $t \to \infty$, and the rate of convergence in practice might be extremely slow. Hence, even when training for, e.g., $10^6$ iterations, gradient descent might reach a solution which is still too far from the KKT point, and therefore reconstruction might fail. Thus, even when training until the gradient of the empirical loss is extremely small, the direction of the network’s parameters might be far from the direction of a KKT point. In \cite{moroshko2020implicit}, the authors proved that in \emph{diagonal linear networks} (i.e., a certain simplified architecture of deep linear networks) the initialization scale controls the rate of convergence to the KKT point, namely, when the initialization is small gradient flow converges much faster to a KKT point.
A similar phenomenon seems to occur also in our empirical results: when training without weight decay, small initialization seems to be required to allow reconstruction. However, when training with weight decay, our theoretical analysis in \cref{sec:general loss theory} explains why small initialization is no longer required. Here, the reconstruction does not rely on converging to a KKT point of the max-magin problem, but relies on \cref{eq:reconstruction eq general loss} which holds (approximately) whenever we reach a sufficiently small gradient of the training objective. Thus, when training with weight decay and reaching a small gradient \cref{eq:reconstruction eq general loss} holds, which allows for reconstruction, contrary to training without weight decay where reaching a small gradient does not imply converging close to a KKT point.

\begin{figure}[htbp]
    \includegraphics[width=\textwidth]{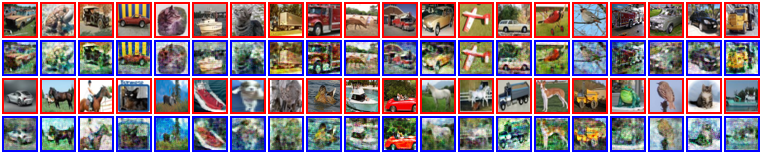}
    \caption{\textbf{Reconstruction from CNN.} Training samples ({\color{red}red}) and their best reconstructions ({\color{blue}blue})}
    \label{fig:convolutions}
    \centering
\end{figure}

\paragraph{Reconstruction from Convolutional Neural Networks (CNNs).} CNNs adhere to the assumptions of \cref{thm:known KKT}, 
yet \cite{haim2022reconstructing} failed to apply their reconstruction scheme~\cref{eq:binary_rec_loss} to CNNs.
We observe that incorporating weight-decay during training (using standard initialization) enables samples reconstruction. In~\cref{fig:convolutions} we show an example for reconstruction from a binary classifier whose first layer is a Conv-layer with kernel size $3$ and $32$ output channels, followed by two fully connected layers ({\sc Conv}($k$=$3$,$C_\text{out}$=$32$)-$1000$-$1$). The weight-decay term is $\lambda_{\text{WD}}$=$0.001$ (the training setup is similar to that of MLP). In~\cref{fig:weight_decay}c we show the reconstructability for the same CNN model trained with other values of $\lambda_{\text{WD}}$. Note how the weight-decay term plays similar role in the CNN as in the MLP case. See full details in~\cref{sec:app_conv}.

\subsection{The Effect of the Number of Parameters and Samples on Reconstructability}
\label{sec:p_n}

\cite{haim2022reconstructing} observed that models trained on fewer samples are more susceptible to reconstruction in terms of both quantity and quality. In this section, we delve deeper into this phenomenon, focusing on the intricate relationship between the number of parameters in the trained model and the number of training samples. We conduct the following experiment:

We train $3$-layer MLPs with architecture $D$-$W$-$W$-1 on $N$ training samples from binary CIFAR10 (animals vs. vehicles), where $W\in\{5, 10, 50, 100, 500, 1000\}$ and $N\in \{10, 50, 100, 300, 500\}$. We conduct the experiment for both classification and regression losses, with BCE and MSE loss respectively. Generalization error is $23\%$-$31\%$ for BCE (classification) and $0.69$-$0.88$ for MSE (regression), compared to $50\%$/$0.97$ for similar models with random weights.

\begin{figure}
    \begin{tabular}{cc}
         \includegraphics[width=.5\textwidth]{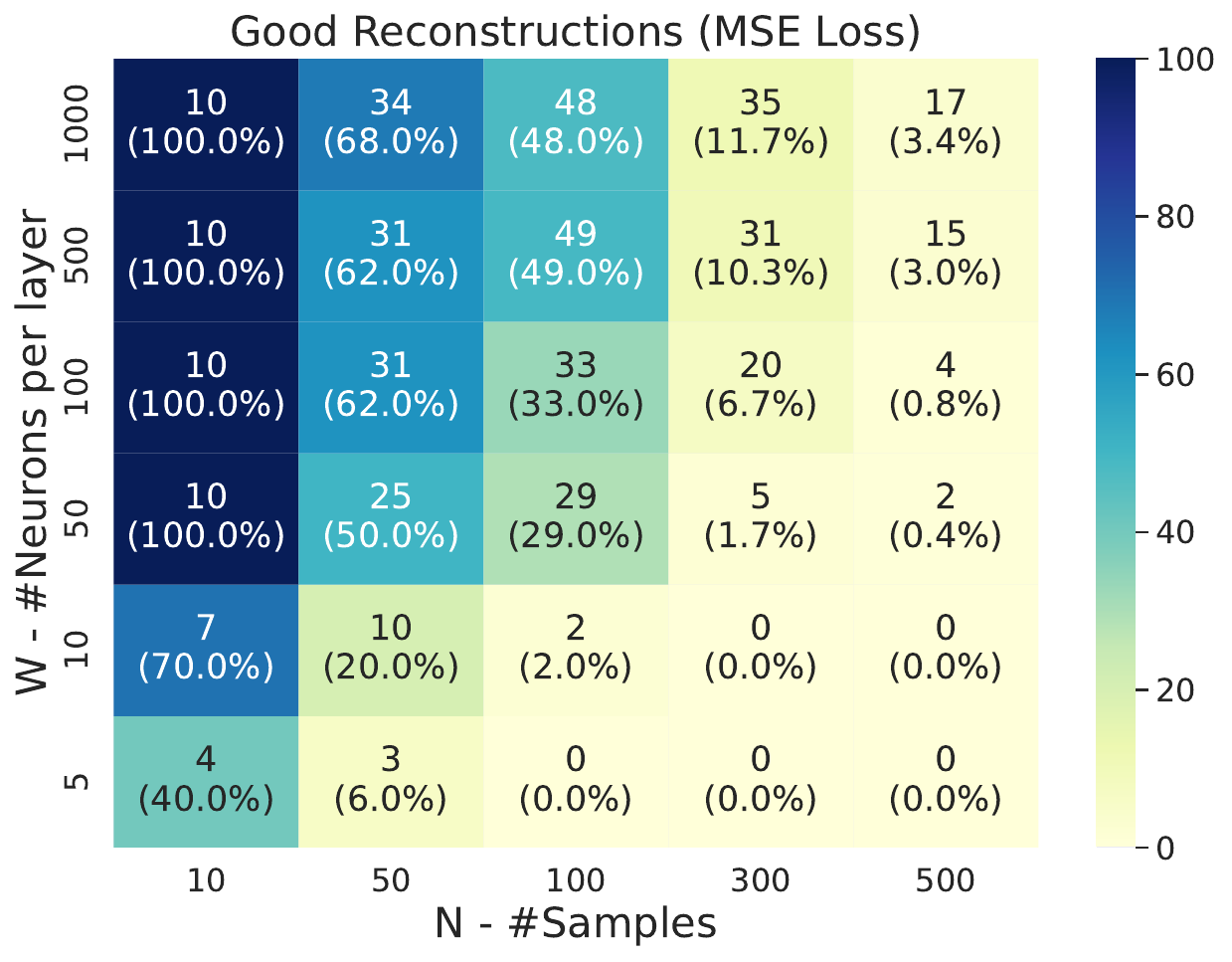}  &
         \includegraphics[width=.5\textwidth]{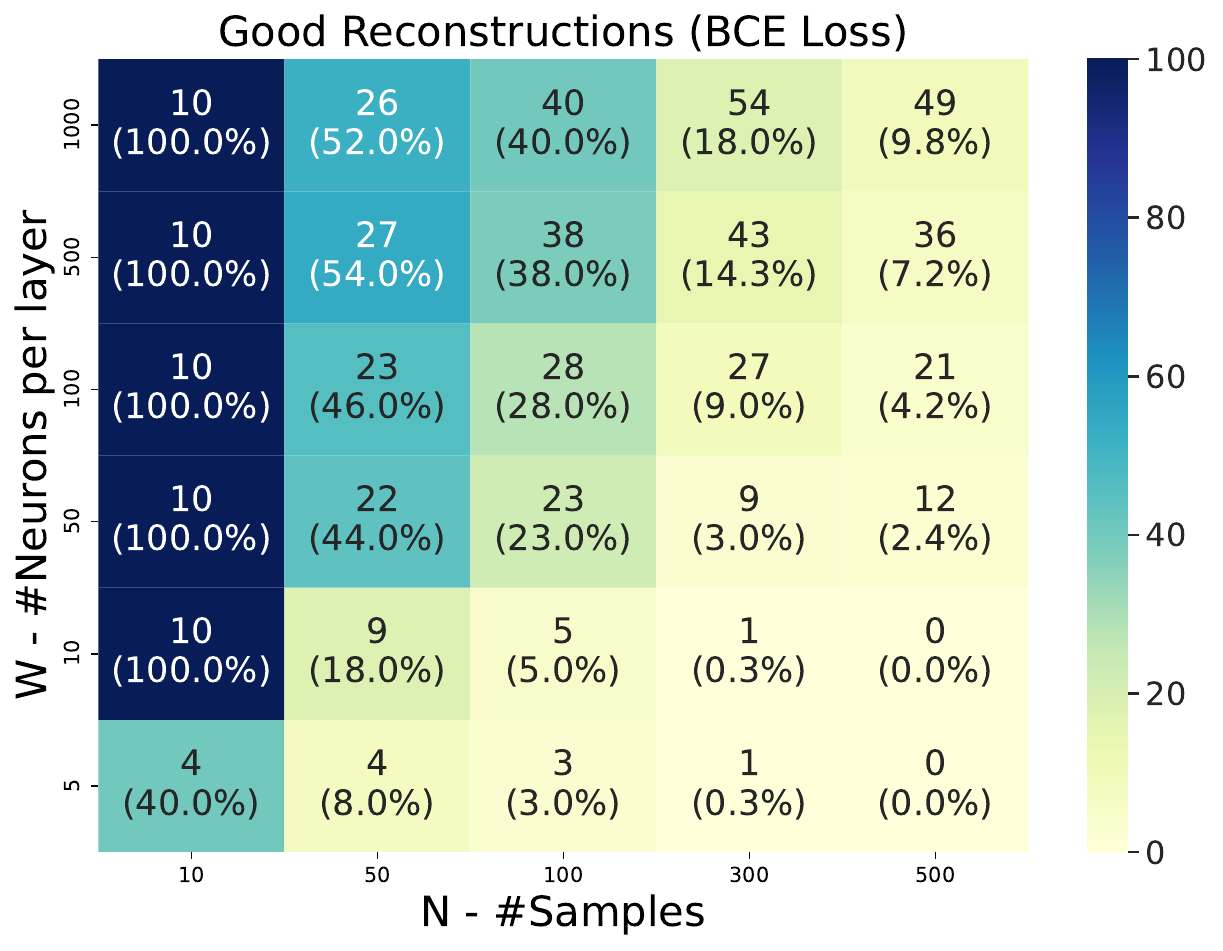} 
    \end{tabular}
    \caption{
    \textbf{Effect of the number of neurons and number of training samples on reconstructability.} We train $3$-layer MLPs with architecture $D$-$W$-$W$-1 on $N$ training samples from binary CIFAR10 (animals vs. vehicles), using MSE (\textit{left}) or BCE (\textit{right}) loss. At each cell we report the number of good reconstructions ($\text{SSIM}$>$0.4$), in both absolute numbers and as a percentage relative to $N$.
    \vspace{-8pt}
    }
\label{fig:pn_effect}
    \centering
\end{figure}

We reconstruct each model using~\cref{eq:general_loss} and record the number of good reconstructions. The results are shown in~\cref{fig:pn_effect}. Note that as $\nicefrac{W}{N}$ increases, our reconstruction scheme is capable of reconstructing more samples, and vice versa. 
For example, consider the case when $N$=$10$. To successfully reconstruct the entire training set, it is sufficient for $W$ to be greater than $50$/$10$ (for MSE/BCE). However, when $N$=$500$, even larger models (with larger $W$) can only reconstruct up to $8\%$ of the samples.

Lastly, we reconstruct from a model with $W$=$10$,$000$, trained on $N$=$5$,$000$ samples ($5$ times larger than any previous model). While there is some degradation in the quality of the reconstructions compared to models trained on fewer samples, it is evident that our scheme can still reconstruct some of the training samples. For full results see \cref{appen: large models many samples}.

\vspace{-4pt}
\section{Conclusions}
\vspace{-4pt}
We present improved reconstruction methods and conduct a comprehensive analysis of the reconstruction method proposed by~\cite{haim2022reconstructing}. Particularly, we extend their reconstruction scheme to a multi-class setting and devise a novel reconstruction scheme for general loss functions, allowing reconstruction in a regression setting (e.g., MSE loss). 
We examine various factors influencing reconstructability. We shed light on the role of weight decay in samples memorization, allowing for sample reconstruction from convolutional neural networks. Lastly, we examine the intricate relationship between the number of parameters, the number of samples, and the vulnerability of the model to reconstruction schemes. 
We acknowledge that our reconstruction method raises concerns regarding privacy. We consider it crucial to present such methodologies as they encourage researchers to study the potential hazards associated with training neural networks. Additionally, it allows for the development of protective measures aimed at preventing the leakage of sensitive information.


\paragraph{Limitations.} While we have relaxed several of the previous assumptions presented in \cite{haim2022reconstructing}, our method still exhibits certain limitations. First, we only consider relatively small-scale models: up to several layers, without residual connections, that have trained for many iterations on a relatively small dataset without augmentations. Second, determining the optimal hyperparameters for the reconstruction scheme poses a challenge as it requires exploring many configurations for even a single model. 

\paragraph{Future work.} All of the above extend our knowledge and understanding of how memorization works in certain neural networks. This opens up several possibilities for future research including extending our reconstruction scheme to practical models (e.g., ResNets), exploring reconstruction from models trained on larger datasets or different data types (e.g., text, time-series, tabular data), analyzing the impact of optimization methods and architectural choices on reconstructability, and developing privacy schemes to protect vulnerable samples from reconstruction attacks.

\paragraph*{Acknowledgements.}
This project received funding from the European Research Council (ERC) under the European Union’s Horizon 2020 research and innovation programme (grant agreement No 788535), and ERC grant 754705, and from the D. Dan and Betty Kahn Foundation.
GV acknowledges the support of the NSF and the Simons Foundation for the Collaboration on the Theoretical Foundations of Deep Learning.

\bibliography{main}
\bibliographystyle{abbrvnat}

\clearpage

\appendix

\section{Analyzing and Visualizing the Results of the Reconstruction Optimization}

The analysis of the results of the various reconstruction losses~\cref{eq:binary_rec_loss,eq:multiclass_rec_loss,eq:general_loss}, involve verifying and checking which of the training samples were reconstructed. 
In this section we provide further details on our method for analyzing the reconstruction results, and how we measure the quality of our reconstructions.

\subsection{Analyzing the Results of the Reconstruction Optimization}
\label{sec:app_analysis_details}

In order to match between samples from the training set and the outputs of the reconstruction algorithm (the so-called "candidates") we follow the same protocol of \cite{haim2022reconstructing}. Note that before training our models, we subtract the mean image from the given training set. Therefore the training samples are $d$-dimensional objects where each entry is in $[-1,1]$.

First, for each training sample we compute the distance to all the candidates using a normalized $L_2$ score:
\begin{equation}
    \label{eq:normalized_l2}
    d(\bx,\by) = \left\Vert \frac{\bx-\mu_\bx}{\sigma_\bx} - \frac{\by-\mu_{\by}}{\sigma_\by} \right\Vert_2^2
\end{equation}

Where $\bx,\by\in \reals^d$ are a training sample or an output candidate from the reconstruction algorithm, $\mu_\bx = \frac{1}{d} \sum_{i=1}^d \bx(i)$ is the mean of $\bx$ and $\sigma_\bx = \sqrt{ \frac{1}{d-1} \sum_{i=1}^d (\bx(i)-\mu_\bx)^2} $ is the standard deviation of $\bx$ (and the same goes for $\by,\mu_\by,\sigma_\by$).

Second, for each training sample, we take $C$ candidates with the smallest distance according to ~\cref{eq:normalized_l2}. $C$ is determined by finding the first candidate whose distance is larger than $B$ times the distance to the closest nearest neighbour (where $B$ is a hyperparameter). Namely, for a training sample $\bx$, the nearest neighbour is $\by_1$ with a distance $d(\bx,\by_1)$, then $C$ is determined by finding a candidate $\by_{C+1}$ whose distance is $d(\bx,\by_{C+1})>B \cdot d(\bx,\by_1)$, and for all $j\leq C$, $d(\bx,\by_j) \leq B \cdot d(\bx,\by_1)$. $B$ was chosen heuristically to be $B=1.1$ for MLPs, and $B=1.5$ for convolutional models. 
The $C$ candidates are then summed to create the reconstructed sample $\hat{\bx} = 
\frac{1}{C} \sum_{j=1}^C \by_j$. In general, we can also take only $C=1$ candidate, namely just one nearest neighbour per training sample, but choosing more candidates improve the visual quality of the reconstructed samples.

Third, the reconstructed sample $\hat{\bx}$ is scaled to an image in $[0,1]$ by adding the training set mean and linearly "stretching" the minimal and maximal values of the result to $[0,1]$.
Finally, we compute the SSIM between the training sample $\bx$ and the reconstructed sample $\hat{\bx}$ to measure the quality of reconstruction.

\subsection{Deciding whether a Reconstruction is ``Good''}
\label{sec:ssim_0.4}

Here we justify our selection for SSIM$=$$0.4$ as the threshold for what we consider as a ``good" reconstruction. In general, the problem of deciding whether a reconstruction is the correct match to a given sample, or whether a reconstruction is a ``good" reconstruction is equivalent to the problem of comparing between images. No ``synthetic" metric (like SSIM, $l2$ etc.) will be aligned with human perception. A common metric for this purpose is LPIPS~\citep{zhang2018perceptual} that uses a classifier trained on Imagenet~\citep{deng2009imagenet}, but since CIFAR images are much smaller than Imagenet images ($32\times 32$ vs. $224\times 224$) it is not clear that this metric will be better than SSIM. 

\begin{figure}
    \centering
    \begin{tabular}{c}
         \includegraphics[width=\textwidth]{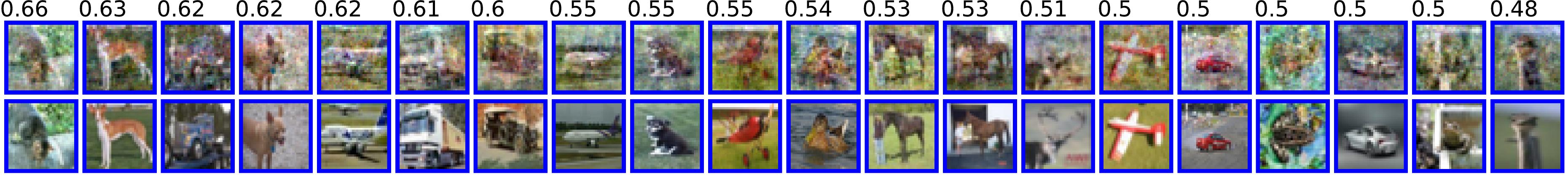}  \\
         \includegraphics[width=\textwidth]{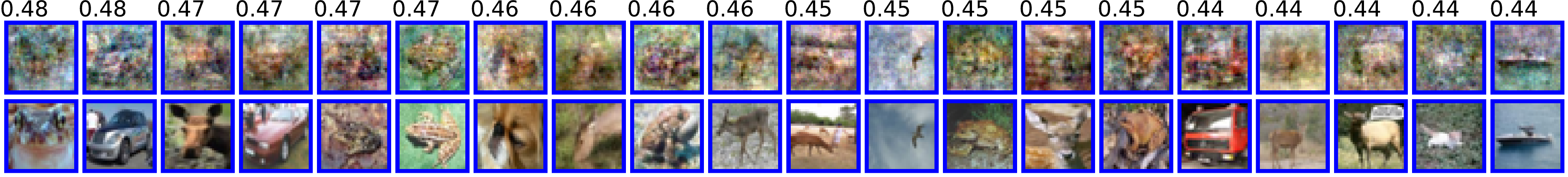}  \\
         \includegraphics[width=\textwidth]{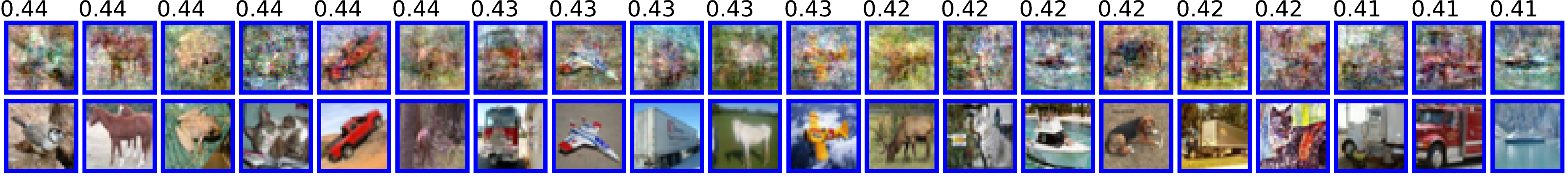}  \\
         \includegraphics[width=\textwidth]{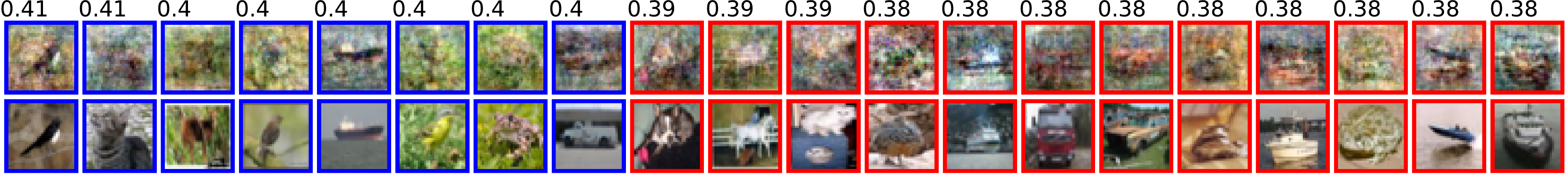}  \\
         \includegraphics[width=\textwidth]{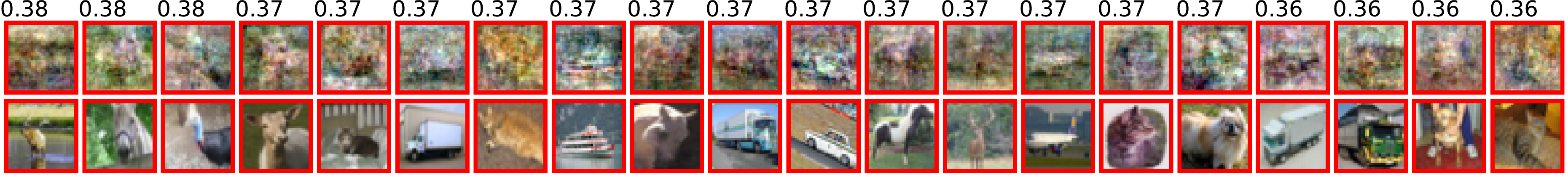}  \\
         \includegraphics[width=\textwidth]{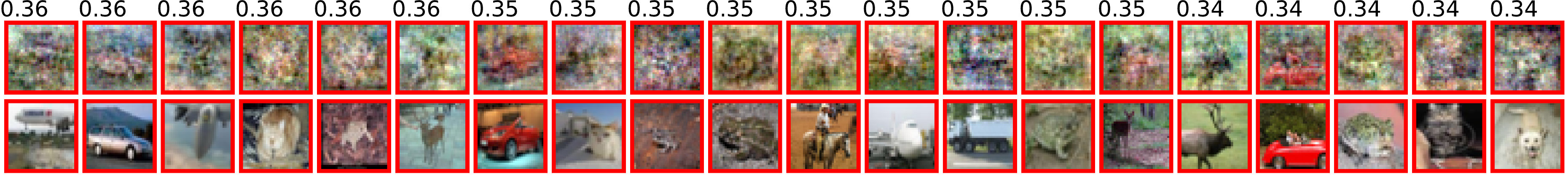} 
    \end{tabular}
    \caption{Justifying the threshold of SSIM$=0.4$ as good rule-of-thumb for a threshold for a ``good" reconstruction.  The SSIM values are shown above each train-reconstruction pair. Note that samples with SSIM$>0.4$ ({\color{blue}blue}) are visually similar. Also some of the samples with SSIM$<0.4$ ({\color{red}red}) are similar. In general deciding whether a reconstruction is ``good" is an open question beyond the scope of this paper.}
    \label{fig:ssim_0.4}
\end{figure}

As a simple rule of thumb, we use SSIM$>$$0.4$ for deciding that a given reconstruction is ``good". To justify, we plot the best reconstructions (in terms of SSIM) in \cref{fig:ssim_0.4}. Note that almost all samples with SSIM$>$$0.4$ are also visually similar (for a human). Also note that some of the samples with SSIM$<$$0.4$ are visually similar, so in this sense we are ``missing" some good reconstructions. In general, determining whether a candidate output of a reconstruction algorithm is a match to a training sample is an open question and a problem in all other works for data reconstruction, see for example~\cite{carlini2023extracting} that derived a heuristic for reconstructed samples from a generative model. This cannot be dealt in the scope of this paper, and is an interesting future direction for our work.

\section{Implementation Details}
\label{sec:app_implemetation}

\paragraph{Further Training Details.} The models that were reconstructed in the main part of the paper were trained with learning rates of $0.01$ for binary classifiers (both MLP and convolutional), and $0.5$ in the case of multi-class classifier (\cref{sec:multiclass}). The models were trained with full batch gradient descent for $10^6$ epochs, to guarantee convergence to a KKT point of~\cref{eq:optimization problem} or a local minima of~\cref{eq:general_loss_with_wd}. When small initialization of the first layer is used (e.g., in~\cref{fig:binary_vs_multiclass,fig:c_vs_dpc}), the weights are initialized with a scale of $10^{-4}$. We note that \cite{haim2022reconstructing} observed that models trained with SGD can also be reconstructed. The experiment in~\cref{appen: large models many samples} (large models with many samples) also uses SGD and results with similar conclusion, that some models trained with SGD can be reconstructed. In general, exploring reconstruction from models trained with SGD is an interesting direction for future works.

\paragraph{Runtime and Hardware.}
Runtime of a single reconstruction run (specific choice of hyperparameters) from a model $D$-$1000$-$1000$-$1$ takes about $20$ minutes on a GPU Tesla V-100 $32$GB or NVIDIA Ampere Tesla A40 $48$GB.

\paragraph{Hyperparameters of the Reconstruction Algorithm.}
Note that the reconstruction loss contains the derivative of a model with ReLU layers, which is flat and not-continuous. Thus, taking the derivative of the reconstruction loss results in a zero function. To address this issue we follow a solution presented in ~\cite{haim2022reconstructing}. Namely, given a trained model, we replace in the backward phase of backpropogation the ReLU function with the derivative of a softplus function (or SmoothReLU) $f(x) = \alpha \log(1 + e^{-x})$, where $\alpha$ is a hyperparameter of the reconstruction scheme. The functionality of the model itself does not change, as in the foraward phase the function remains a ReLU.
Only the backward function is replaced with a smoother version of the derivative of ReLU which is $f'(x)=\alpha \sigma(x) = \frac{\alpha}{1+e^{-x}}$ (here $\sigma$ is the Sigmoid function). 
To find good reconstructions we run the algorithm multiple times (typically $100$ times) with random search over the hyperparameters (using the Weights \& Biases framework~\citep{wandb}). The exact parameters for the hyperparameters search are:

\begin{itemize}
    \item Learning rate: log-uniform in $[10^{-5},1]$
    \item $\sigma_x$: log-uniform in $[10^{-6},0.1]$
    \item $\lambda_\text{min}$: uniform in $[0.01,0.5]$
    \item $\alpha$: uniform in $[10,500]$
\end{itemize}

\vspace{-8pt}
\section{Multiclass Reconstruction - More Results}
\vspace{-8pt}
\label{sec:fixed_dataset_size}

\subsection{Experiments with Different Number of Classes and Fixed Training Set Size}

\begin{figure}[htbp]
    \centering
    \includegraphics[width=\textwidth]{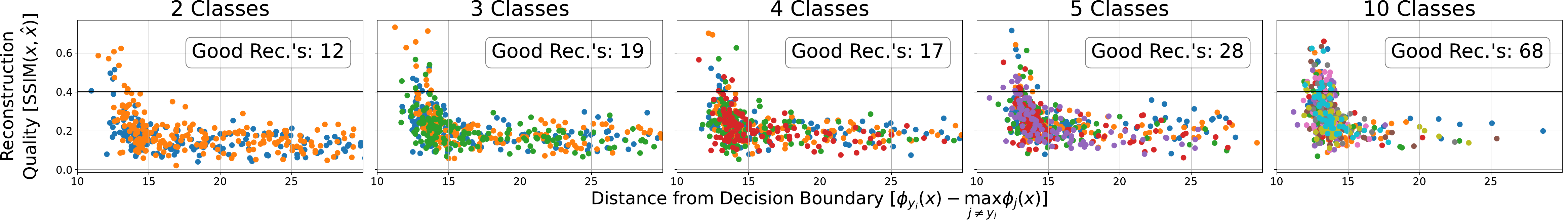}
    \caption{Experiments of reconstruction from models trained on a a fixed training set size ($500$ samples) for different number of classes. Number of ``good" reconstruction is shown for each model.}
    \label{fig:c_vs_dpc_500}\vspace{-8pt}
\end{figure}

To complete the experiment shown in \cref{fig:c_vs_dpc}, we also perform experiments on models trained on various number of classes ($C \in \{2,3,4,5,10\}$) and with a fixed training set size of $500$ samples (distributed equally between classes), see \cref{fig:c_vs_dpc_500}. 
It can be seen that as the number of classes increases, also does the number of good reconstructions, where for $10$ classes there are more than $6$ times good reconstructions than for $2$ classes. Also, the quality of the reconstructions improves as the number of classes increase, which is 
depicted by an overall higher SSIM score.
We also note, that the number of good reconstructions in \cref{fig:c_vs_dpc_500} is very similar to the number of good reconstructions from \cref{fig:c_vs_dpc} for $50$ samples per class. We hypothesize that although the number of training samples increases, the number of "support vectors" (i.e samples on the margin which can be reconstructed) that are required for successfully interpolating the entire dataset does not change by much.

\vspace{-8pt}
\subsection{Results on SVHN Dataset}
\begin{figure}[!hbp]
    \centering
    \includegraphics[width=.9\textwidth]{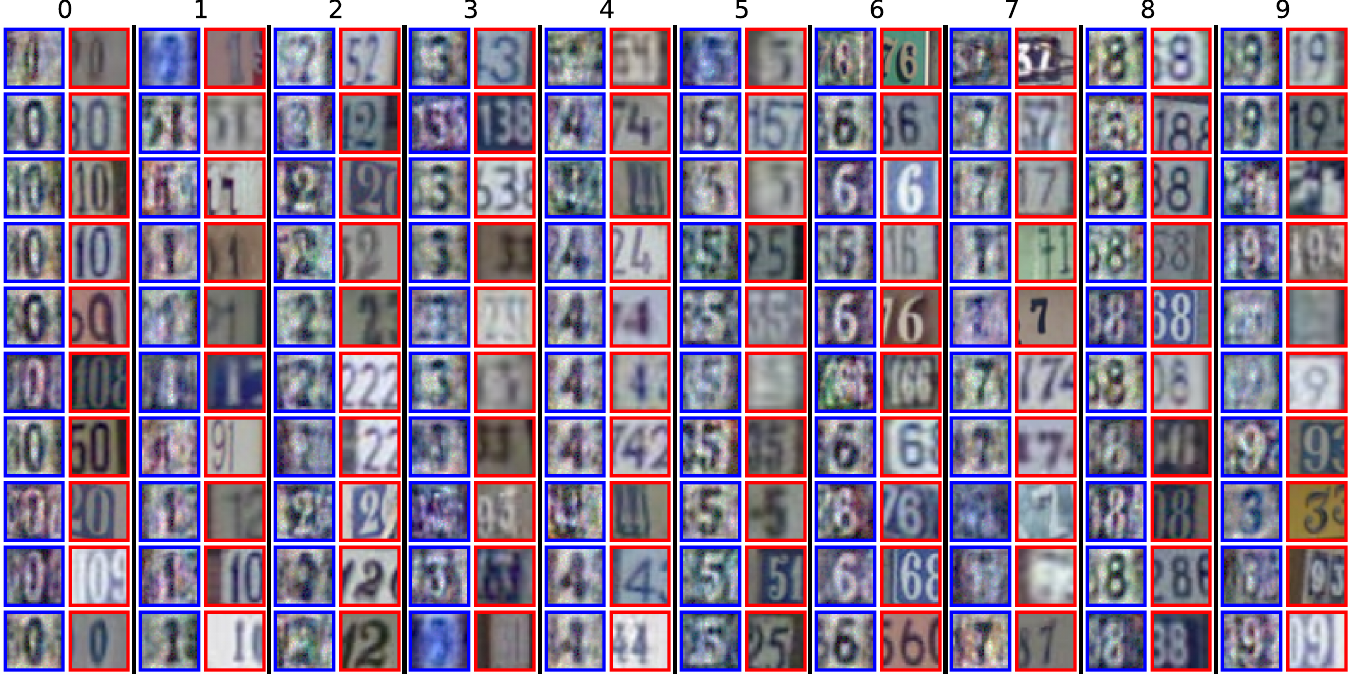}
    \caption{Reconstruction form model trained on $50$ samples per class from the SVHN dataset.}
    \label{fig:svhn}
\end{figure}

\vspace{-8pt}
As shown in~\cref{fig:svhn}, our multiclass reconstruction scheme is not limited to CIFAR10 dataset, but can be extended to other datasets as well, specifically SVHN dataset~\citep{netzer2011reading}. The model whose reconstructions are shown in~\cref{fig:svhn} was trained on $50$ samples per class (total of $10$ classes) and the rest of the training hyperparameters are the same as that of its CIFAR10 equivalent (of $50$ sample per class with $10$ classes).

\newpage
\section{General Losses - More Results}
\label{sec:app_general_losses}

\begin{figure}[!htp]
     \includegraphics[width=\textwidth]{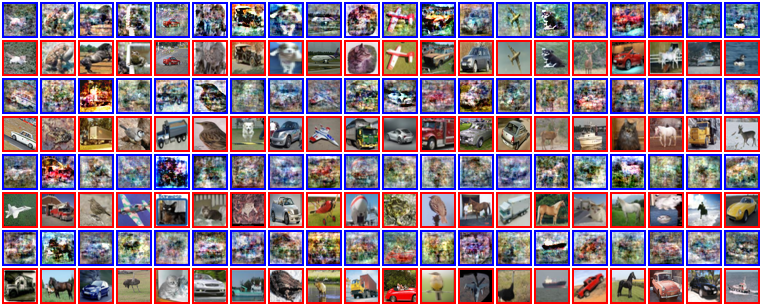} \\
     \caption{\textbf{Reconstruction using $L_2$ loss.} Training samples ({\color{red}red}) and their best reconstructions ({\color{blue}blue}) using an MLP classifier that was trained on 300 CIFAR10 images using an $L_2$ regression loss, as described in~\cref{sec:general_losses} and~\cref{fig:general_losses}.} 
\label{fig:general_losses_samples_l2}
\centering
\end{figure}

\begin{figure}[!htp]
     \includegraphics[width=\textwidth]{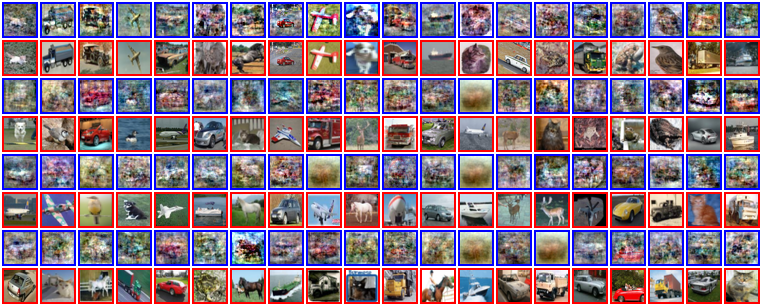} \\
     \caption{\textbf{Reconstruction using $L_{2.5}$ loss.} Training samples ({\color{red}red}) and their best reconstructions ({\color{blue}blue}) using an MLP classifier that was trained on 300 CIFAR10 images using an $L_{2.5}$ regression loss, as described in~\cref{sec:general_losses} and~\cref{fig:general_losses}.} 
\label{fig:general_losses_samples_l2.5}
\centering
\end{figure}

\begin{figure}[!htp]
     \includegraphics[width=\textwidth]{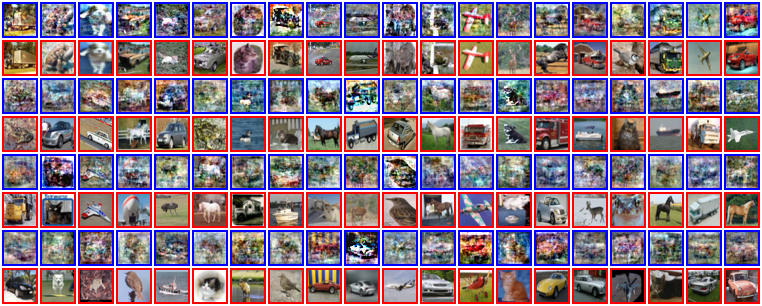} \\
     \caption{\textbf{Reconstruction using Huber loss.} Training samples ({\color{red}red}) and their best reconstructions ({\color{blue}blue}) using an MLP classifier that was trained on 300 CIFAR10 images using Huber loss, as described in~\cref{sec:general_losses} and~\cref{fig:general_losses}.} 
\label{fig:general_losses_samples_lhuber}
\centering
\end{figure}

Following the discussion in \cref{sec:general_losses} and~\cref{fig:general_losses}, Figures \ref{fig:general_losses_samples_l2}, \ref{fig:general_losses_samples_l2.5}, \ref{fig:general_losses_samples_lhuber} present visualizations of training samples and their reconstructions from models trained with $L_2$, $L_{2.5}$ and Huber loss, respectively.

\newpage

\section{Further Analysis of Weight Decay}
\label{sec:app_weight_decay}

By looking at the exact distribution of reconstruction quality to the distance from the margin, we observe that weight-decay (for some values) results in more training samples being on the margin of the trained classifier, thus being more vulnerable to our reconstruction scheme. 

This observation is shown in~\cref{fig:weight_decay_scatter} where we show the scatter plots for all the experiments from~\cref{fig:weight_decay} (a). We also provide the train and test errors for each model. It seems that the test error does not change significantly. However, an interesting observation is that reconstruction is possible even for models with non-zero training errors, i.e. models that do not interpolate the data, for which the assumptions of~\cite{lyu2019gradient} do not hold.

\noindent%
\begin{minipage}{\linewidth}%
\makebox[\linewidth]{%
  \includegraphics[keepaspectratio=true,width=\textwidth]{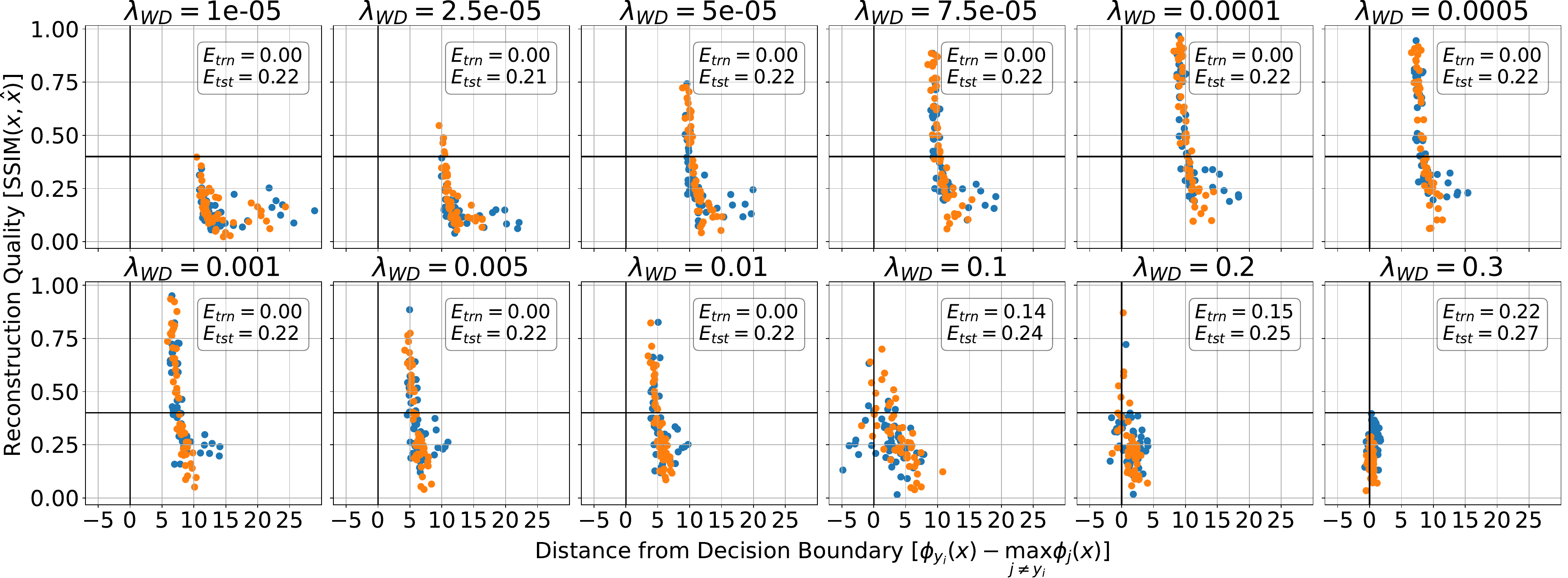}}
\captionof{figure}{Scatter plots of the $12$ experiments from \cref{fig:weight_decay} (a). Each plot is  model trained with a different value of weight decay on $2$ classes with $50$ samples in each class. Certain values of weight decay make the model more susceptible to our reconstruction scheme.
}\label{fig:weight_decay_scatter}%
\end{minipage}

\section{Convolutional Neural Networks - Ablations and Observations}
\label{sec:app_conv}

In this section we provide more results and visualizations to the experiments on convolutional neural network in~\cref{sec:weight_decay}.

\begin{figure}[htbp]
     \includegraphics[width=\textwidth]{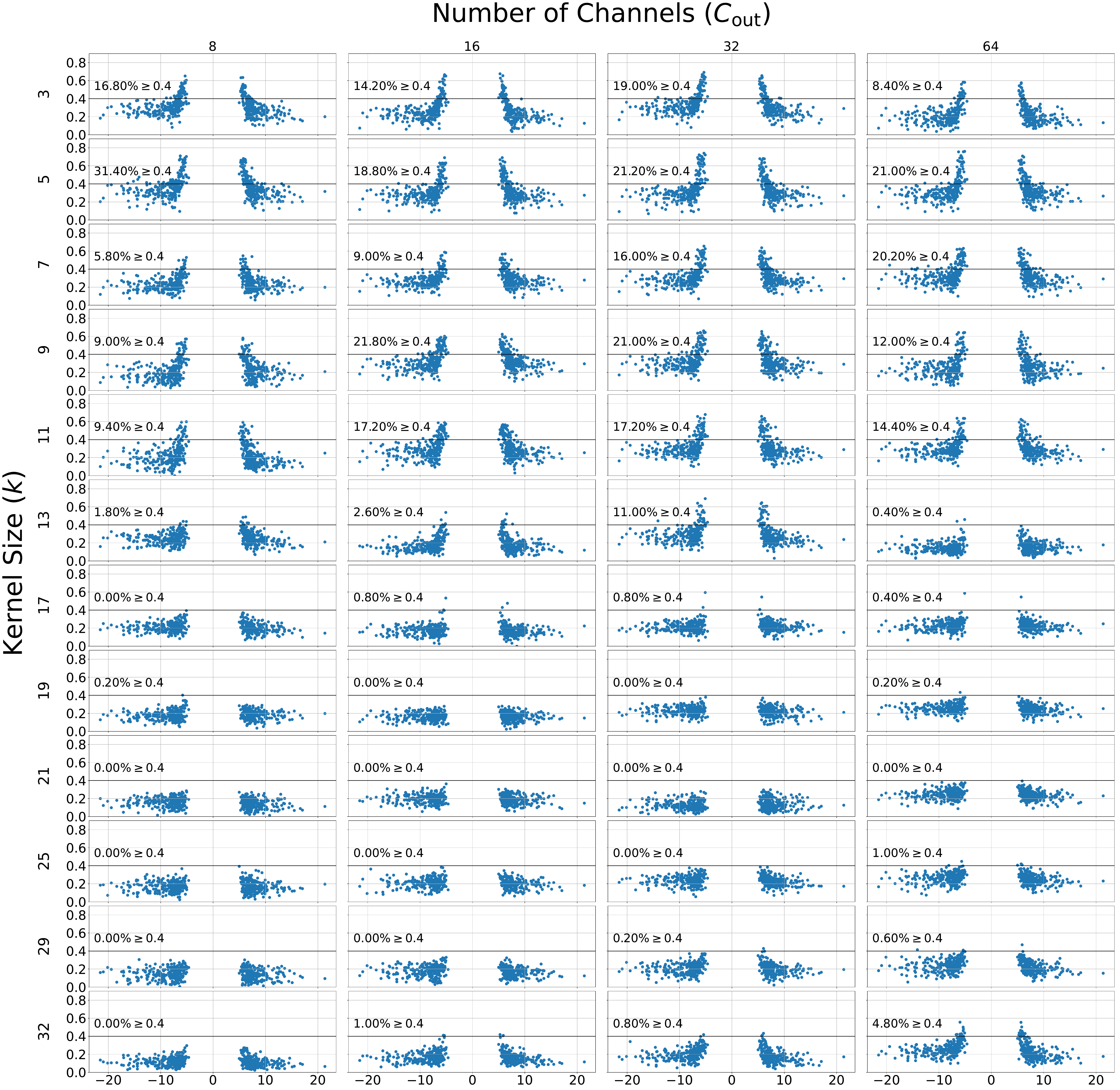}
    \caption{\textbf{Ablating the choice of the kernel size and output-channels} for reconstruction from neural binary classifiers with architecture {\sc Conv}(kernel-size=$k$,output-channels=$C_\text{out}$)-$1000$-$1$. (Please note that this figure might not be displayed well on Google Chrome. Please open in Acrobat reader.)}
    \label{fig:conv_grid}
    \centering
\end{figure}

In~\cref{fig:conv_grid} we show ablations for the choice of the kernel-size ($k$) and number of output channels ($C_\text{out}$) for models with architecture {\sc Conv}(kernel-size=$k$,output-channels=$C_\text{out}$)-$1000$-$1$. All models were trained on $500$ images ($250$ images per class) from the CIFAR10 dataset, with weight-decay term $\lambda_{\text{WD}}$=$0.001$. As can be seen, for such convolutional models we are able to reconstruct samples for a wide range of choices.

Note that the full summary of reconstruction quality versus the distance from the decision boundary for the model whose reconstucted samples are shown in~\cref{fig:convolutions}, is shown in~\cref{fig:conv_grid} for kernel-size $3$ (first row) and number of output channels $32$ (third column).

\paragraph{Further analysis of~\cref{fig:conv_grid}.}
As expected for models with less parameters, the reconstructability decreases as the number of output channels decrease. An interesting phenomenon is observed for varying the kernel size: for a fixed number of output channel, as the kernel size increases, the susceptibility of the model to our reconstruction scheme decreases.
However, as the kernel size approaches $32$ (the full resolution of the input image), the reconstructability increases once again. On the one hand it is expected, since for kernel-size=$32$ the model is essentially an MLP, albeit with smaller hidden dimension than usual (at most $64$ here, whereas the typical model used in the paper had $1000$). On the other hand, it is not clear why for some intermediate values of kernel size (in between $3$ and $32$) the reconstructability decreases dramatically (for many models there are no reconstructed samples at all). This observation is an interesting research direction for future works.

\begin{figure}[htbp]
    \includegraphics[width=\textwidth]{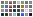}
    \caption{The kernels of the model whose reconstructions are shown in \cref{fig:convolutions}, displayed as RGB images.}
    \label{fig:conv_first_layer}
\end{figure}

\newpage
\paragraph{Visualizing Kernels.} In~\cite{haim2022reconstructing}, it was shown that some of the training samples can be found in the first layer of the trained MLPs, by reshaping and visualizing the weights of the first fully-connected layer. As opposed to MLPs, in the case of a model whose first layer is a convolution layer, this is not possible. For completeness, in~\cref{fig:conv_first_layer} we visualize all $32$ kernels of the Conv layer. Obviously, full images of shape $3$x$32$x$32$ cannot be found in kernels of shape $3$x$3$x$3$, which makes reconstruction from such models (with convolution first layer) even more interesting.

\section{Reconstruction From a Larger Number of Samples}\label{appen: large models many samples}

One of the major limitations of \cite{haim2022reconstructing} is that they reconstruct from models that trained on a relatively small number of samples. Specifically, in their largest experiment, a model is trained with only $1$,$000$ samples. Here we take a step further, and apply our reconstruction scheme for a model trained on $5$,$000$ data samples. 

To this end, we trained a 3-layer MLP, where the number of neurons in each hidden layer is $10$,$000$.  Note that the size of the hidden layer is $10$ times larger than in any other model we used. Increasing the number of neurons seems to be one of the major reasons for which we are able to reconstruct from such large datasets, although we believe it could be done with smaller models, which we leave for future research. 
We used the CIFAR100 dataset, with 50 samples in each class, for a total of $5000$ samples.

\begin{figure}[ht]
    \begin{tabular}{c}
         \includegraphics[width=\textwidth]{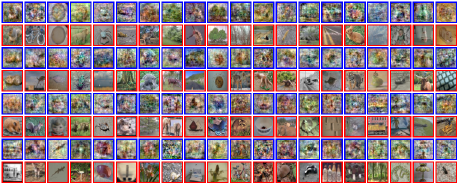} \\
         (a) Full Images. Original samples from the training set (\textcolor{red}{\textit{red}}) and reconstructed results (\textcolor{blue}{\textit{blue}})\\
         \includegraphics[width=\textwidth]{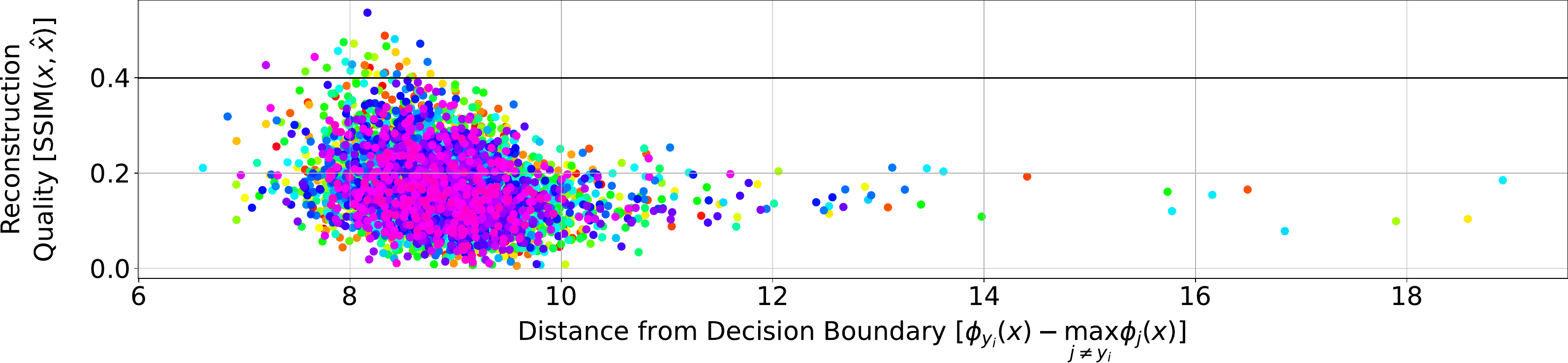} \\
         (b) Scatter plot (similar to \cref{fig:c_vs_dpc}) . 
    \end{tabular}
    \caption{Reconstruction from a model trained on $50$ images per class from the CIFAR100 dataset ($100$ classes, total of $5000$ datapoints). The model is a $3$-layer MLP with $10000$ neurons in each layer. (Please note that this figure might not be displayed well on Google Chrome. Please open in Acrobat reader.)} 
    \label{fig:cifar_100_scatter}
    \centering
\end{figure}

In \cref{fig:cifar_100_scatter}a we give the best reconstructions of the model. Note that although there is a degradation in the quality of the reconstruction w.r.t a model trained on less samples, it is still clear that our scheme can reconstruct some of the training samples to some extent. In \cref{fig:cifar_100_scatter}b we show a scatter plot of the SSIM score w.r.t the distance from the boundary, similar to \cref{fig:c_vs_dpc}a. Although most of the samples are on or close to the margin, only a few dozens achieve an SSIM$>0.4$. This may indicate that there is a potential for much more images to reconstruct, and possibly with better quality.

%


\end{document}